\documentclass[11pt]{article}
\PassOptionsToPackage{table}{xcolor}

\usepackage[preprint]{acl} 
\usepackage{times}          
\usepackage{latexsym}       
\usepackage[T1]{fontenc}    
\usepackage[utf8]{inputenc} 
\usepackage{microtype}      
\usepackage{inconsolata}    

\usepackage{amsmath}
\usepackage{amsfonts}
\usepackage{amssymb}        

\usepackage{graphicx}
\usepackage{booktabs}       
\usepackage{tabularx}       
\usepackage{array}
\usepackage{multirow}       
\usepackage{caption}        
\usepackage{subcaption}     
\usepackage[table]{xcolor}  

\usepackage[linesnumbered,ruled,vlined]{algorithm2e}

\usepackage{tcolorbox}      

\newcommand{\annotate}[3]{\textcolor{#2}{#1}}

\title{ScRPO: From Errors to Insights}

\author{
    Lianrui Li$^{\diamondsuit}$\thanks{Equal contribution.}~,
    Dakuan Lu$^{\diamondsuit}$\footnotemark[1]~,
    Jiawei Shao$^{\diamondsuit}$~,
    \textbf{Xuelong Li}$^{\diamondsuit}$\thanks{Corresponding author.}\\
    $^\diamondsuit$Institute of Artificial Intelligence (TeleAI), China Telecom \\
    \fontsize{10.2pt}{0.1\baselineskip}\selectfont \texttt{lilianrui338@gmail.com}
}

\begin{document}
\maketitle
\begin{abstract}
We introduce \textbf{S}elf-\textbf{c}orrection \textbf{R}elative \textbf{P}olicy \textbf{O}ptimization~(ScRPO), a novel reinforcement learning framework designed to empower large language models with advanced mathematical reasoning capabilities through iterative self-reflection and error correction. The ScRPO framework operates in two distinct phases: (1) Trial-and-error learning stage, where the model is trained via GRPO, and incorrect responses are collected to form an ``error pool''; and (2) Self-correction learning stage, which guides the model to introspectively analyze and rectify the reasoning flaws behind its previous errors. Extensive evaluations across challenging mathematical benchmarks, including AIME, AMC, Olympiad, MATH-500, and GSM8k, validate the efficacy of our approach. Using DeepSeek-R1-Distill-Qwen-1.5B and 7B as backbones, ScRPO achieves average accuracies of 64.8\% and 77.8\%, respectively. This represents a significant improvement of 6.0\% and 3.2\% over vanilla baselines, consistently outperforming strong post-training methods such as DAPO and GRPO. These findings establish ScRPO as a robust paradigm for enabling autonomous self-improvement in AI systems, particularly in tasks with limited external feedback.

\end{abstract}

\begin{figure}[h]
    \centering
    \includegraphics[width=0.8\linewidth]{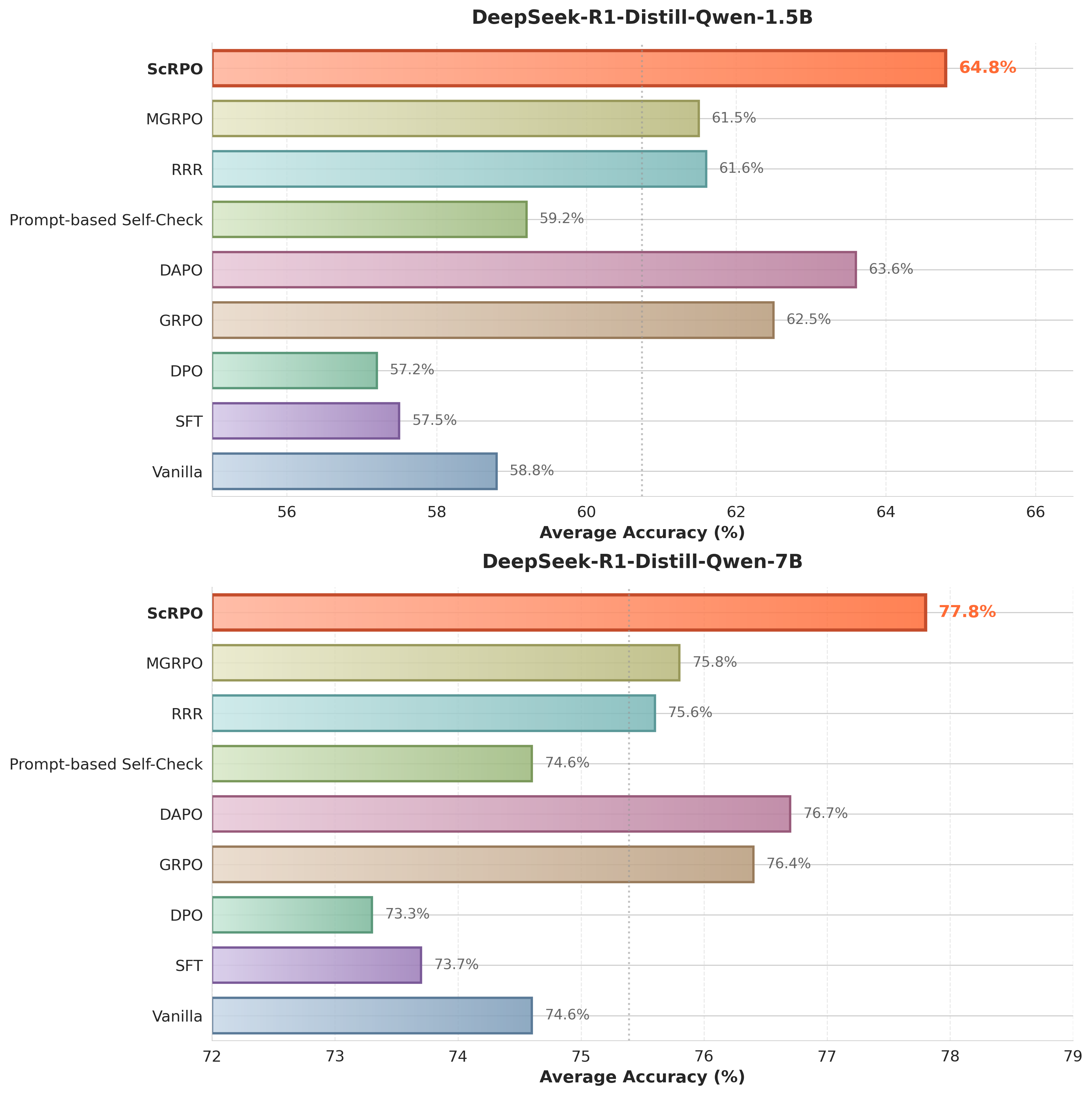}
    \caption{Average accuracy of DeepSeek-R1-Distill-Qwen-1.5B and 7B models across various benchmarks. ScRPO achieves the highest performance across both model sizes.}
    \label{fig:intro fig}
\end{figure}

\section{Introduction}
Large language models (LLMs) have demonstrated remarkable capabilities across diverse domains~\citep{Zhao2025, matarazzo2025surveylargelanguagemodels,  shen2025satorireinforcementlearningchainofactionthought}, yet they continue to struggle with complex mathematical reasoning tasks that require multi-step logical inference and precise computational skills~\citep{anand2024mathifyevaluatinglargelanguage, loureiro2025advancingmultistepmathematicalreasoning}. Recent advances in reinforcement learning from human feedback (RLHF)~\citep{RLHF} and its variants have shown promise in improving model performance on such challenging problems~\citep{kaufmann2024surveyreinforcementlearninghuman, an2025aiflowperspectivesscenarios}. However, conventional reinforcement learning methods such as Group Relative Policy Optimization (GRPO)~\citep{shao2024deepseekmathpushinglimitsmathematical}, while effective at enhancing performance through reward maximization, face several fundamental limitations~\citep{sane2025hybridgrouprelativepolicy}. First, these methods extract only limited information from scalar rewards, failing to capture the nuanced reasons behind errors~\citep{liu2025structuralrewardmodelenhancing}.
Second, they struggle to effectively utilize high-difficulty problems for parameter updates, as unsuccessful attempts provide minimal learning signal. Third, they typically process each training sample only once, resulting in low data efficiency and limited opportunity for iterative improvement~\citep{chen2023maybe05dataneeded}. These limitations motivate the need for mechanisms that enable models to systematically analyze their errors and learn from mistakes—a capability that is central to human mathematical learning.

\begin{figure*}
    \includegraphics[width=1\linewidth]{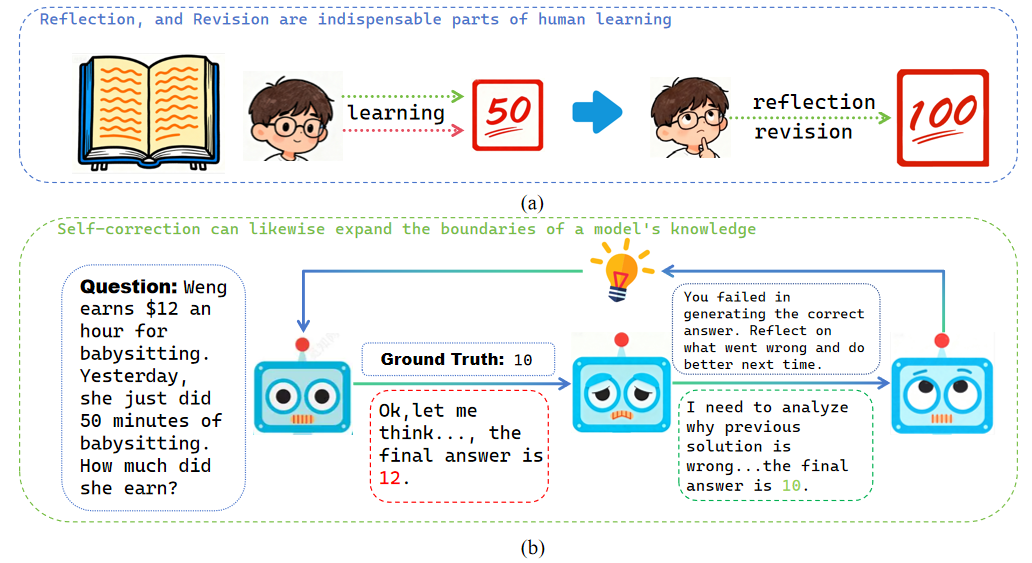}
    \caption{Illustration of the motivation behind ScRPO. (a) \textbf{How humans learn mathematics.} In the process of learning math, solving new problems is not enough; reflecting on and summarizing lessons from previous errors is a crucial step. (b) \textbf{ScRPO Paradigm.} When an LLM solves a math problem, it may encounter failure. By explicitly analyzing and learning from these errors, the model's reasoning capability is expanded.}
    \label{fig:humanvsgpt}
\end{figure*}

Human mathematical learning operates through a fundamentally different paradigm centered on error recognition, reflective analysis, and systematic correction, shown in Figure~\ref{fig:humanvsgpt}(a). Human learners naturally engage in metacognitive reasoning when faced with mathematical challenges: they deconstruct erroneous solutions, diagnose underlying conceptual misunderstandings, and formulate targeted strategies for future problem solving~\citep{didolkar2024metacognitivecapabilitiesllmsexploration}.
This reflective learning framework not only improves computational accuracy but also develops transferable reasoning skills that are generalized effectively to novel mathematical contexts.

Inspired by this human learning paradigm, we propose Self-correction Relative Policy Optimization~(ScRPO), shown in Figure~\ref{fig:humanvsgpt}(b), a novel reinforcement learning framework that enhances LLMs' mathematical reasoning capabilities through systematic self-reflection and error correction. Unlike traditional approaches that focus solely on reward optimization, ScRPO explicitly incorporates a mistake-driven learning mechanism that enables models to learn systematically from their failures.

Our main contributions are as follows. We introduce ScRPO, a novel framework that integrates self-reflection and error correction mechanisms into reinforcement learning for mathematical reasoning. To demonstrate the efficacy of our approach, we conducted extensive experiments on the training dataset and evaluated performance against several established benchmarks. Empirical results indicate that ScRPO significantly outperforms existing baselines, particularly in complex tasks requiring multi-step verification and iterative refinement. 

\section{Preliminaries}
\label{sec:preliminary}
In this section, we provide the necessary background on Group Relative Policy Optimization (GRPO)~\citep{shao2024deepseekmathpushinglimitsmathematical}. For each query $q$, GRPO samples a group of outputs $\{o_1, \ldots, o_G\}$ from the policy $\pi_\theta$. The advantage $\hat{A}_i$ for each output is computed relative to the group's distribution:
\begin{equation}
    \hat{A}_i = \frac{r(q, o_i) - \bar{r}}{\sigma}, \quad \text{where } \bar{r} = \frac{1}{G}\sum_{j=1}^G r(q, o_j)
\end{equation}
Here, $\sigma$ is the standard deviation of the group rewards. To ensure stable updates, GRPO maximizes the following surrogate objective:
\begin{equation}
\label{eq:grpo_simple_loss}
\begin{split}
    \mathcal{J}(\theta) = \mathbb{E} \bigg[ &\min\left( \rho_t \hat{A}_i, \text{clip}(\rho_t, 1-\epsilon, 1+\epsilon)\hat{A}_i \right) \\
    - \beta \mathbb{D}_{KL} \bigg]
\end{split}
\end{equation}
where $\rho_t = \frac{\pi_\theta(o_t | q, o_{<t})}{\pi_{old}(o_t | q, o_{<t})}$ is the importance ratio, $\epsilon$ is the clipping threshold, and $\beta$ controls the KL divergence penalty $\mathbb{D}_{KL}$ relative to the reference policy. We provide the detailed token-level derivation in Appendix~\ref{app:grpo_details}.

\textbf{Training Process.} During training, for each batch of prompts, GRPO: (1) samples multiple responses per prompt from the current policy, (2) evaluates rewards for all responses, (3) computes group-relative advantages, and (4) updates the policy parameters to maximize the expected advantage while respecting the clipping and KL constraints.

Our proposed ScRPO method builds upon this GRPO framework by introducing self-reflection and error correction mechanisms, which we detail in the following section.

\section{Methodology}
In this section, we present our ScRPO training framework for enhancing reasoning capabilities in LLMs. Our approach consists of two core stages: the trial-and-error learning stage and the self-correction learning stage. We begin by providing an overview of the overall framework architecture. In the following subsections, we describe the key components of ScRPO, including the error collection mechanism, the self-correction process, and the reward attribution strategy.

\subsection{Framework Overview}

\begin{figure*}
    \centering
    \includegraphics[width=1\linewidth]{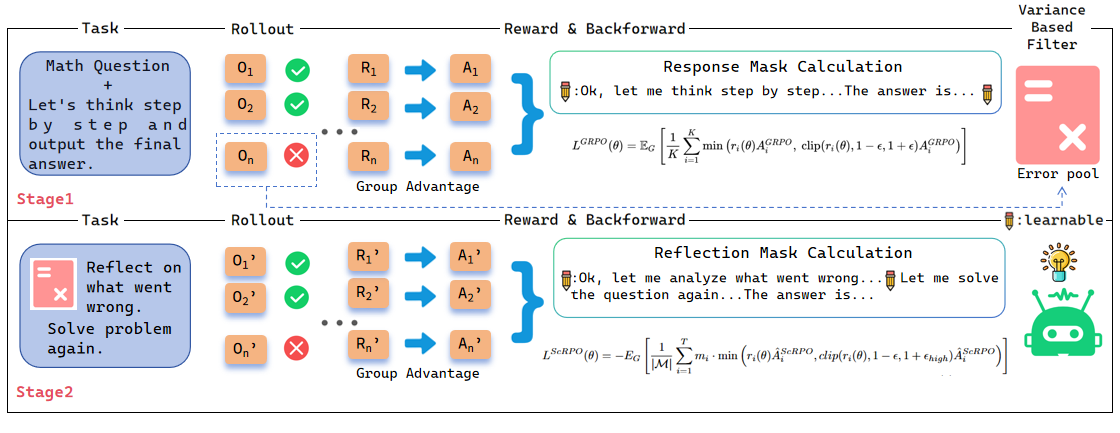}
    \caption{An illustration of the proposed ScRPO method. ScRPO consists
 of continuous trial-and-error learning that collects incorrect answers, followed by periodic self-correction learning, where the model reflects on errors and generates improved responses.}
    \label{fig:main}
\end{figure*}

The detailed methodology of our ScRPO training framework is illustrated in Figure~\ref{fig:main}. ScRPO comprises two sequential stages that work in tandem to improve model reasoning through reflection and correction.

The first stage, the trial-and-error learning stage, follows a paradigm similar to conventional GRPO. Building upon the standard GRPO framework, we incorporate a module called Variance-Based Filter, which is designed to eliminate questions that are either excessively challenging or overly trivial. During this stage, we dynamically collect instances where the model produces erroneous responses, maintaining a repository of incorrect question-answer pairs throughout the training process. This repository (error pool) serves as the foundation for the subsequent self-correction stage.

The second stage, the self-correction learning stage, is activated periodically (every 5 training iterations, supported by ablation studies in Appendix~\ref{app:ablation_f}). During this phase, the model is required to perform reflective analysis on the original questions and their associated incorrect responses, subsequently generating revised answers informed by the erroneous outputs. When the model produces a correct response after reflection, the corresponding reward signal is attributed to the token sequences associated with the reflection component, thereby reinforcing the self-correction behavior. The complete pseudo-code of the overall algorithm is provided in Appendix~\ref{app:scrpo-algorithm}.

\subsection{Variance-Based Filter}
In mathematical reasoning reinforcement learning scenarios, we propose a Variance-Based Filter (VBF) to enhance ScRPO training by selectively utilizing informative negative samples. The core principle is to identify problems with intermediate difficulty—those where the model exhibits partial understanding but inconsistent performance.

For each problem prompt $p$, we sample $n$ responses (e.g., $n=12$) from the current policy $\pi_\theta$. Let $\mathcal{R}_p = \{O_1, O_2, \ldots, O_n\}$ denote the set of sampled responses for problem $p$, and define the correctness indicator:
\begin{equation}
c_i = \begin{cases}
1 & \text{if response } O_i \text{ is correct} \\[0.5em]
0 & \text{otherwise}
\end{cases}
\end{equation}
The empirical accuracy for problem $p$ is computed as:
\begin{equation}
\begin{gathered}
\text{acc}(p) = \frac{1}{n}\sum_{i=1}^{n} c_i
\end{gathered}
\end{equation}
The problem $p$ is retained if its accuracy falls within the variance-rich region:
\begin{equation}
\begin{gathered}
p \in \mathcal{P}_{\text{filter}} \iff \text{acc}(p) \in (0.33, 0.66)
\end{gathered}
\end{equation}
This range indicates that the problem exhibits high behavioral variance—neither trivially easy (acc $\approx$ 1) nor impossibly hard (acc $\approx$ 0) for the current policy. Such problems are most informative for learning as they represent the model's knowledge boundary.

The variance of a Bernoulli-distributed outcome (correct/incorrect) is given by:
\begin{equation}
\begin{gathered}
\text{Var}(p) = \text{acc}(p) \cdot (1 - \text{acc}(p))
\end{gathered}
\end{equation}
This variance is maximized when $\text{acc}(p) = 0.5$, achieving $\text{Var}_{\max} = 0.25$. Within our selected range $(0.33, 0.66)$, the minimum variance is:
$$\text{Var}_{\min} = 0.33 \times 0.66 = 0.2222$$

High variance problems provide maximum information gain, as the model's predictions are most uncertain and sensitive to policy improvements in this region.

By focusing training on these high-variance negative samples, we enable the model to:
\begin{itemize}
    \item Learn from mistakes on problems within its capability range
    \item Avoid overfitting to trivial problems or being overwhelmed by impossible ones
    \item Concentrate gradient updates on the most informative decision boundaries
\end{itemize}

\subsection{Self-correction learning stage}

In the self-correction learning stage, we construct an error pool using the question-answer pairs and incorrect solutions extracted by the Variance-Based Filter. We construct a new prompt using the template shown in Figure~\ref{fig:critique-prompt}.

\begin{figure}[t]
\centering
\begin{tcolorbox}[
  width=0.9\linewidth,
  colframe=blue!70!black,
  colback=blue!5!white,
  arc=4mm,
  boxrule=0.8pt
]
You tried performing the task, but failed to generate the correct answer. Reflect on what went wrong and do better next time.

\textbf{Question:} \texttt{\{question\}}

\textbf{Wrong solution:} \texttt{\{wrong\_answer\}}

\textbf{**Analysis:**} [Analyze why the solution is wrong]

\textbf{**Corrected Solution:**} [Provide the correct solution]

\textbf{Response:}
\end{tcolorbox}
\caption{Prompt template for self-correction learning stage.}
\label{fig:critique-prompt}
\end{figure}

\textbf{Success-Conditioned Gradient Attribution Mechanism.} 
We structure the newly generated prompts into the standard input data format of VeRL~\citep{Sheng_2025}. The error pool is fed back into the policy model for multiple sampling rounds ($G$ samples per prompt), and we compute rewards and advantages based on the correctness of the final generated answers. 

Crucially, while the reward signal $r_j \in \{0,1\}$ is determined by whether the **final corrected answer** is correct, we employ a **selective gradient attribution** strategy: gradient updates are applied exclusively to the reflection tokens (the "Analysis" section). This is achieved by redefining the response mask $m_{j,i}$ to identify all tokens between \texttt{**Analysis:**} and \texttt{**Corrected Solution:**} using string matching, and setting only these tokens' mask values to 1.

The advantage $\hat{A}_j$ for each trajectory $j$ is computed using the group-relative method:
\begin{equation}
\hat{A}_j = r_j - \frac{1}{G}\sum_{k=1}^{G} r_k
\end{equation}
where $r_j$ reflects the correctness of the final answer. However, during backpropagation, this advantage is only attributed to masked reflection tokens ($m_{j,i}=1$).

\textbf{Design Rationale.} This approach enables the model to learn \emph{what kind of reflective analysis leads to successful corrections}, rather than directly optimizing answer generation or being constrained to specific task patterns. By attributing success-conditioned rewards exclusively to the self-correction process, the model develops transferable capabilities to diagnose errors and formulate effective correction strategies. The loss function in the self-correction learning stage can be expressed as follows:

\begin{equation}
\resizebox{\linewidth}{!}{$\displaystyle
\begin{aligned}
    \mathcal{L}_{\text{ScRPO}}(\theta) 
    &= -\mathbb{E}_{\substack{q \sim P(Q), \\ \{o_j\}_{j=1}^{G} \sim \pi_{\theta_{old}}(O|q)}} 
    \Bigg[ \frac{1}{G} \sum_{j=1}^{G} \frac{1}{|\mathcal{M}_j|} \sum_{i=1}^{T_j} m_{j,i} \cdot \\
    &\quad \Bigg( \min \bigg( 
        \frac{\pi_\theta(a_{j,i} | s_{j,i}, q)}{\pi_{\theta_{old}}(a_{j,i} | s_{j,i}, q)} \hat{A}_{j,i}, \\
    &\quad \text{clip} \left( 
        \frac{\pi_\theta(a_{j,i} | s_{j,i}, q)}{\pi_{\theta_{old}}(a_{j,i} | s_{j,i}, q)}, 1 - \epsilon_{\text{low}}, 1 + \epsilon_{\text{high}} 
    \right) \hat{A}_{j,i} 
    \bigg) \\
    &\quad - \beta \mathbb{D}_{KL} \left[ \pi_{\theta}(a_{j,i}|s_{j,i}, q) \parallel \pi_{ref}(a_{j,i}|s_{j,i}, q) \right] \Bigg) \Bigg]
\end{aligned}$}
\end{equation}

\text{where:}

\begin{equation}
m_{j,i} = \begin{cases}
1 & \text{if token } i \text{ in trajectory } j \in \{\text{Analysis}\} \\
0 & \text{otherwise}
\end{cases}
\end{equation}

\begin{equation}
\resizebox{0.95\linewidth}{!}{$\displaystyle
|\mathcal{M}_j| = \sum_{i=1}^{T_j} m_{j,i} \quad \text{(number of selected tokens in trajectory } j\text{)}
$}
\end{equation}

\begin{equation}
\resizebox{0.95\linewidth}{!}{$\displaystyle
G = \text{group size (number of trajectories sampled per query)}
$}
\end{equation}

\begin{equation}
T_j = \text{length of trajectory } j
\end{equation}

\section{Experiments}
\subsection{Experiment Setup}
\textbf{Benchmarks.} We evaluate mathematical reasoning on five benchmarks covering elementary to olympiad-level difficulty, including GSM8k~\citep{cobbe2021trainingverifierssolvemath}, MATH-500~\citep{lightman2023letsverifystepstep}, AIME-2024, AMC (10/12), and Olympiad~\citep{he2024olympiadbenchchallengingbenchmarkpromoting}. We report accuracy (Acc) for all datasets.

\textbf{Implementation Details.} Training is conducted on a mixed-difficulty dataset combining MATH and DAPO-MATH~\citep{han2025modelsthoughtenoughtraining}, after removing all Chinese-language problems, resulting in approximately 14k examples with no overlap with evaluation benchmarks. Detailed training configurations, hyperparameters, and ablation studies are provided in Appendices~\ref{sec:benchmarks_and_training_details} and ~\ref{sec:appendix_ablation}.

\subsection{Baselines and Models}
We evaluate ScRPO against standard baselines, including SFT, DPO, GRPO, and DAPO, on five mathematical reasoning benchmarks. The results for SFT and DPO are taken from~\citep{han2025modelsthoughtenoughtraining}, as we use the same datasets and models as in their work. In addition to these traditional methods, we introduce two closely related approaches as baselines: \textbf{Multi-layer GRPO (MGRPO)}~\citep{ding2025multilayergrpoenhancingreasoning} and \textbf{Reflect, Retry, Reward (RRR)}~\citep{bensal2025reflectretryrewardselfimproving}. Although MGRPO employs a two-stage GRPO framework, it performs secondary verification on all problems rather than focusing specifically on self-correction of errors. Moreover, MGRPO does not incorporate variance filtering or targeted gradient updates at specific token positions. While RRR proposes self-reflection mechanisms, it primarily focuses on self-correction rather than adopting a two-stage training paradigm where the model first learns and then corrects errors. Furthermore, we introduce a complementary comparison methods: \textbf{Prompt-based Self-Check}, which performs inference-time self-correction without parameter updates. For models, two model variants, DeepSeek-R1-Distill-Qwen-1.5B and DeepSeek-R1-Distill-Qwen-7B, are evaluated under identical training configurations. 

\subsection{Main Results}
Table~\ref{table:comparison} presents the experimental results of ScRPO and baseline methods across five mathematical reasoning benchmarks. We summarize the key findings as follows:

\textbf{ScRPO consistently outperforms all baselines.} 
Across both model scales, ScRPO achieves the highest average accuracy, reaching 64.8\% on the 1.5B model and 77.8\% on the 7B model. Compared to the vanilla baseline, ScRPO yields substantial improvements of +6.0\% and +3.2\% on average, respectively. Notably, ScRPO surpasses strong reinforcement learning baselines such as GRPO (+2.3\% and +1.4\%) and DAPO (+1.2\% and +1.1\%), demonstrating the effectiveness of our proposed self-correction mechanism.

\textbf{Traditional fine-tuning methods show limited effectiveness.} 
Surprisingly, both SFT and DPO lead to performance degradation on challenging benchmarks. For the 1.5B model, SFT results in a -5.9\% drop on AMC and -4.6\% on Olympiad, while DPO exhibits similar trends with -4.5\% on AIME24. These results suggest that naive supervised learning or preference optimization may not effectively elicit the reasoning capabilities of distilled models, potentially due to distribution shift or overfitting issues.

\textbf{ScRPO excels on competition-level problems.} 
The performance gains of ScRPO are particularly pronounced on more challenging benchmarks. On AIME24, ScRPO achieves improvements of +5.7\% (1.5B) and +4.7\% (7B), substantially outperforming other reflection-based methods such as RRR (+2.8\% and +1.5\%) and MGRPO (+3.1\% and +2.1\%). This indicates that our structured self-correction approach is especially beneficial for complex multi-step reasoning tasks.

\textbf{Prompt-based self-check shows marginal improvements.} 
The prompt-based self-check method yields only modest gains (+0.4\% on average for 1.5B) and even negative results on some benchmarks (-1.8\% on AIME24 for 7B). This highlights the limitation of inference-time prompting strategies and underscores the necessity of training-based approaches like ScRPO to internalize robust self-correction capabilities.

\textbf{Scaling behavior analysis.} 
While the absolute performance improves with model scale, the relative gains from ScRPO remain consistent. This suggests that ScRPO provides complementary benefits that do not diminish as model capacity increases, making it a promising approach for enhancing reasoning abilities across different model sizes.
\begin{table*}[t]
\centering
\footnotesize
\caption{Experimental results on mathematical reasoning benchmarks with several baselines and ScRPO methods, including reflection and error-driven variants, where the "$\Delta$" represents the improvement relative to the "Vanilla" method.}
\label{table:comparison}
\setlength\tabcolsep{8pt} 
\renewcommand{\arraystretch}{1.1}
\begin{tabular}{@{}lccccccccccl@{}} 
\toprule
\multirow{2}{*}{\textbf{Method}} 
& \multicolumn{2}{c}{\textbf{AIME24}} & \multicolumn{2}{c}{\textbf{AMC}} & \multicolumn{2}{c}{\textbf{Olympiad}} & \multicolumn{2}{c}{\textbf{GSM8k}} & \multicolumn{2}{c|}{\textbf{MATH-500}} & \textbf{Avg.} \\
& {Acc} & {$\Delta$} & {Acc} & {$\Delta$} & {Acc} & {$\Delta$} & {Acc} & {$\Delta$} & {Acc} & {$\Delta$} & {Acc} \\ 
\hline
\\[-0.8em]
\multicolumn{12}{l}{{\cellcolor[rgb]{0.957,0.957,0.957}}\textit{\textbf{DeepSeek-R1-Distill-Qwen-1.5B}}} \\
\\[-0.8em]
\textit{Vanilla} & 28.5 & -- & 63.1 & -- & 47.1 & -- & 75.7 & -- & 79.8 & -- & 58.8 \\
\rowcolor[rgb]{0.95,0.95,1.0}
\textit{+SFT} & 27.7 & {\annotate{-0.8}{blue}{}} & 57.2 & {\annotate{-5.9}{blue}{}} & 42.5 & {\annotate{-4.6}{blue}{}} & 81.4 & {\annotate{+5.7}{red}{}} & 78.8 & {\annotate{-1.0}{blue}{}} & 57.5 \\
\rowcolor[rgb]{0.95,0.95,1.0}
\textit{+DPO} & 24.0 & {\annotate{-4.5}{blue}{}} & 59.0 & {\annotate{-4.1}{blue}{}} & 44.0 & {\annotate{-3.1}{blue}{}} & 80.2 & {\annotate{+4.5}{red}{}} & 78.6 & {\annotate{-1.2}{blue}{}} & 57.2 \\
\rowcolor[rgb]{0.95,0.95,1.0}
\textit{+GRPO} & 30.0 & {\annotate{+1.5}{red}{}} & 66.6 & {\annotate{+3.5}{red}{}} & 49.4 & {\annotate{+2.3}{red}{}} & 83.5 & {\annotate{+7.8}{red}{}} & 83.0 & {\annotate{+3.2}{red}{}} & 62.5 \\
\rowcolor[rgb]{0.95,0.95,1.0}
\textit{+DAPO} & 31.5 & {\annotate{+3.0}{red}{}} & 66.8 & {\annotate{+3.7}{red}{}} & 50.6 & {\annotate{+3.5}{red}{}} & 83.7 & {\annotate{+8.0}{red}{}} & 85.3 & {\annotate{+5.5}{red}{}} & 63.6 \\
\rowcolor[rgb]{1.0,0.98,0.94}
\textit{+Prompt-based Self-Check} & 29.2 & \annotate{+0.7}{red}{} & 63.1 & {\annotate{0.0}{blue}{}} & 47.4 & \annotate{+0.3}{red}{} & 77.0 & \annotate{+1.3}{red}{} & 79.4 & {\annotate{-0.4}{blue}{}} & 59.2 \\
\rowcolor[rgb]{0.95,1.0,0.95}
\textit{+RRR} & 31.3 & \annotate{+2.8}{red}{} & 65.3 & {\annotate{+2.2}{red}{}} & 50.1 & \annotate{+3.0}{red}{} & 79.0 & \annotate{+3.3}{red}{} & 82.4 & {\annotate{+2.6}{red}{}} & 61.6 \\
\rowcolor[rgb]{0.95,1.0,0.95}
\textit{+MGRPO} & 31.6 & \annotate{+3.1}{red}{} & 64.9 & {\annotate{+1.8}{red}{}} & 49.7 & \annotate{+2.6}{red}{} & 79.3 & \annotate{+3.6}{red}{} & 82.0 & {\annotate{+2.2}{red}{}} & 61.5 \\
\rowcolor[rgb]{1.0,0.95,0.95}
\textit{\textbf{+ScRPO}} & \textbf{34.2} & {\annotate{+5.7}{red}{}} & \textbf{68.5} & {\annotate{+5.4}{red}{}} & \textbf{52.0} & {\annotate{+4.9}{red}{}} & \textbf{85.0} & {\annotate{+9.3}{red}{}} & \textbf{84.1} & {\annotate{+4.3}{red}{}} & \textbf{64.8} \\
\\[-0.3em]
\hline
\\[-0.8em]
\multicolumn{12}{l}{{\cellcolor[rgb]{0.957,0.957,0.957}}\textit{\textbf{DeepSeek-R1-Distill-Qwen-7B}}} \\
\\[-0.8em]
\textit{Vanilla} & 52.8 & -- & 78.6 & -- & 63.7 & -- & 86.4 & -- & 91.7 & -- & 74.6 \\
\rowcolor[rgb]{0.95,0.95,1.0}
\textit{+SFT} & 48.7 & {\annotate{-4.1}{blue}{}} & 78.6 & {\annotate{0.0}{blue}{}} & 62.7 & {\annotate{-1.0}{blue}{}} & 87.3 & {\annotate{+0.9}{red}{}} & 91.4 & {\annotate{-0.3}{blue}{}} & 73.7 \\
\rowcolor[rgb]{0.95,0.95,1.0}
\textit{+DPO} & 53.0 & {\annotate{+0.2}{red}{}} & 77.2 & {\annotate{-1.4}{blue}{}} & 60.4 & {\annotate{-3.3}{blue}{}} & 86.1 & {\annotate{-0.3}{blue}{}} & 90.0 & {\annotate{-1.7}{blue}{}} & 73.3 \\
\rowcolor[rgb]{0.95,0.95,1.0}
\textit{+GRPO} & 54.0 & {\annotate{+1.2}{red}{}} & 82.1 & {\annotate{+3.5}{red}{}} & 64.0 & {\annotate{+0.3}{red}{}} & 90.2 & {\annotate{+3.8}{red}{}} & 91.9 & {\annotate{+0.2}{red}{}} & 76.4 \\
\rowcolor[rgb]{0.95,0.95,1.0}
\textit{+DAPO} & 54.1 & {\annotate{+1.3}{red}{}} & 82.7 & {\annotate{+4.1}{red}{}} & 64.3 & {\annotate{+0.6}{red}{}} & 90.1 & {\annotate{+3.7}{red}{}} & 92.2 & {\annotate{+0.5}{red}{}} & 76.7 \\
\rowcolor[rgb]{1.0,0.98,0.94}
\textit{+Prompt-based Self-Check} & 51.0 & {\annotate{-1.8}{blue}{}} & 79.0 & {\annotate{+0.4}{red}{}} & 63.4 & {\annotate{-0.3}{blue}{}} & 87.4 & {\annotate{+1.0}{red}{}} & 92.1 & {\annotate{+0.4}{red}{}} & 74.6 \\
\rowcolor[rgb]{0.95,1.0,0.95}
\textit{+RRR} & {54.3} & {\annotate{+1.5}{red}{}} & {80.7} & {\annotate{+2.1}{red}{}} & {63.0} & {\annotate{-0.7}{blue}{}} & {89.1} & {\annotate{+2.7}{red}{}} & {91.0} & {\annotate{-0.7}{blue}{}} & {75.6} \\
\rowcolor[rgb]{0.95,1.0,0.95}
\textit{+MGRPO} & {54.9} & {\annotate{+2.1}{red}{}} & {81.2} & {\annotate{+2.6}{red}{}} & {63.3} & {\annotate{-0.4}{blue}{}} & {88.8} & {\annotate{+2.4}{red}{}} & {90.9} & {\annotate{-0.8}{blue}{}} & {75.8} \\
\rowcolor[rgb]{1.0,0.95,0.95}
\textit{\textbf{+ScRPO}} & \textbf{57.5} & {\annotate{+4.7}{red}{}} & \textbf{83.5} & {\annotate{+4.9}{red}{}} & \textbf{65.1} & {\annotate{+1.4}{red}{}} & \textbf{90.8} & {\annotate{+4.4}{red}{}} & \textbf{92.3} & {\annotate{+0.6}{red}{}} & \textbf{77.8} \\
\bottomrule
\end{tabular}
\end{table*}

\subsection{Self-correction Ability Analysis}
\begin{figure}
    \centering
    \includegraphics[width=1\linewidth]{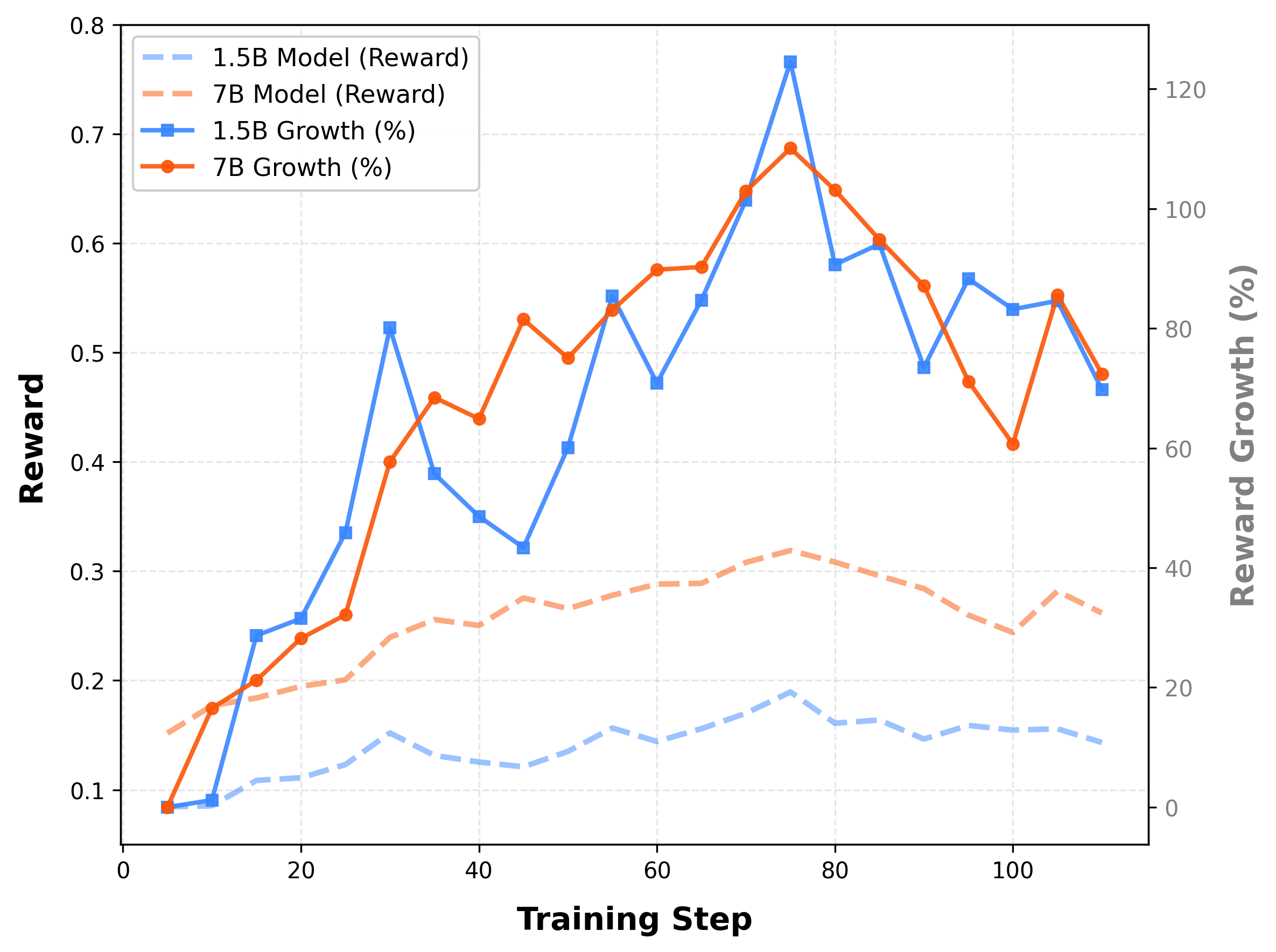}
    \caption{Reward and relative reward growth of DeepSeek-R1-Distill-Qwen-1.5B and DeepSeek-R1-Distill-Qwen-7B during self-correction learning stage}
    \label{fig:reflection reward}
\end{figure}

The reward trajectories during the self-correction learning stage, as illustrated in Figure~\ref{fig:reflection reward}, validate the effectiveness of the ScRPO framework across different model scales. Mathematical reasoning tasks inherently require a progressive, step-by-step chain of inference, where each intermediate reasoning step builds upon previous ones and demands a high level of logical rigor. An error in the early stages of reasoning, or an analysis that deviates substantially from the original problem, typically propagates through subsequent steps and prevents the model from reaching a correct final answer.

Under this setting, the ability to identify and correct erroneous reasoning paths is crucial. Therefore, in our experiments, we adopt the accuracy on the \emph{error pool} as a direct and principled metric for evaluating the model’s self-correction capability, as it explicitly measures whether the model can revise previously incorrect solutions into correct ones through reflection and re-generation.

Both DeepSeek-R1-Distill-Qwen-1.5B and DeepSeek-R1-Distill-Qwen-7B exhibit a clear upward trend in rewards, indicating that the models progressively learn to refine their reasoning through periodic self-correction. Notably, the 7B model consistently outperforms the smaller 1.5B variant, starting from a higher baseline reward ($0.152$ vs. $0.084$) and achieving a substantially higher peak ($0.319$ vs. $0.190$). This performance gap suggests that larger model capacity enables more robust reasoning revision and more effective self-correction, particularly in complex, multi-step mathematical inference scenarios.

\subsection{Ablation Study}
\begin{figure}
    \centering
    \includegraphics[width=0.75\linewidth]{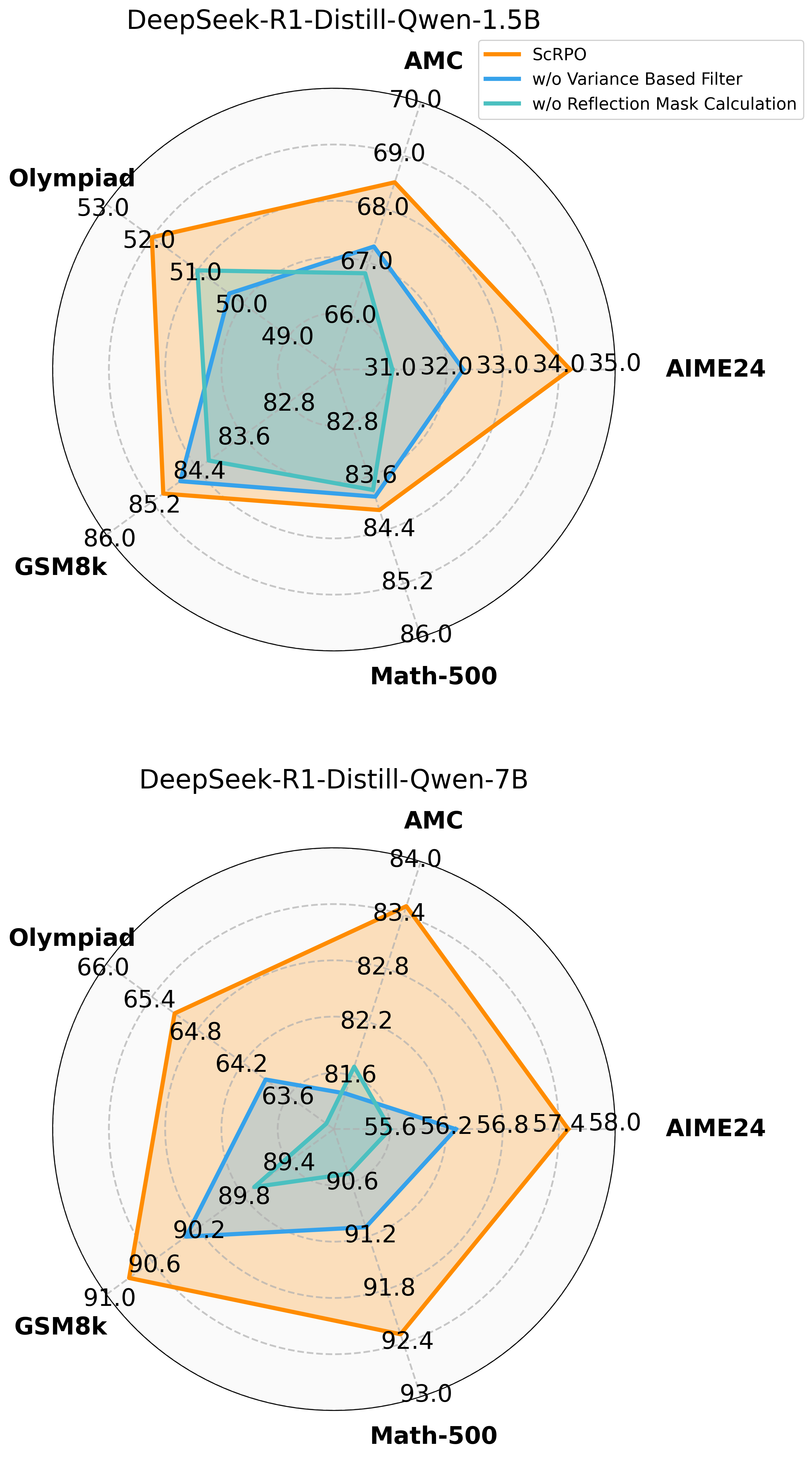}
    \caption{Ablation study results of DeepSeek-R1-Distill-Qwen-1.5B and DeepSeek-R1-Distill-Qwen-7B on various benchmarks}
    \label{fig:circle}
\end{figure}

To verify the effectiveness of the two core components of our proposed ScRPO, we conduct ablation studies on both model sizes, as shown in Table~\ref{tab:ablation} and Figure~\ref{fig:circle}.

\textbf{Effectiveness of Variance-Based Filter.} We analyze ScRPO's performance without variance filtering to assess its contribution. Removing the variance-based filter leads to a performance decrease of 1.1 points for the 1.5B model and 1.2 points for the 7B model compared to the full ScRPO method. These results validate that the variance filtering mechanism not only saves computational resources by focusing on informative samples but also consistently improves model performance across different model scales.

\textbf{Effectiveness of Reflection Mask Calculation.} We further examine the impact of selective loss calculation by comparing full-token loss against reflection-only loss during training. When applying loss to all tokens rather than exclusively to reflection tokens, the 1.5B model's performance drops by 1.4 points, while the 7B model shows a more substantial decline of 1.7 points. This demonstrates the critical importance of targeted loss computation on reflection tokens, as it enables the model to develop robust self-correction capabilities without interference from task-specific answer tokens. The consistent performance improvement across both model scales suggests that proper loss masking is essential for effective reasoning enhancement.

\begin{table}[t]
\centering
\footnotesize
\caption{Ablation study on DeepSeek-R1-Distill-Qwen-1.5B and DeepSeek-R1-Distill-Qwen-7B.}
\label{tab:ablation}
\setlength\tabcolsep{12pt} 
\renewcommand{\arraystretch}{1.3}
\begin{tabular}{@{}lcc@{}} 
\toprule
\textbf{Method} & \textbf{Average} & \textbf{$\Delta$} \\
\midrule
\multicolumn{3}{l}{{\cellcolor[HTML]{F4F4F4}}\textit{\textbf{DeepSeek-R1-Distill-Qwen-1.5B}}} \\
\textit{ScRPO} & \textbf{64.8} & - \\
\textit{w/o Variance-Based Filter} & 63.7 & \textcolor{blue}{-1.1} \\
\textit{w/o Reflection Mask Calculation} & 63.4 & \textcolor{blue}{-1.4} \\
\midrule
\multicolumn{3}{l}{{\cellcolor[HTML]{F4F4F4}}\textit{\textbf{DeepSeek-R1-Distill-Qwen-7B}}} \\
\textit{ScRPO} & \textbf{77.8} & - \\
\textit{w/o Variance-Based Filter} & 76.6 & \textcolor{blue}{-1.2} \\
\textit{w/o Reflection Mask Calculation} & 76.1 & \textcolor{blue}{-1.7} \\
\bottomrule
\end{tabular}
\end{table}

\section{Related Work}
Reinforcement learning has shown substantial potential for improving reasoning capabilities in LLMs, particularly in mathematics and programming tasks~\citep{10884554}. By leveraging feedback signals derived from logical consistency or code execution, these methods encourage models to explore valid reasoning paths more effectively than standard supervised fine-tuning. In this section, we review recent advances in reinforcement learning for reasoning and self-correction mechanisms that enable models to improve their own outputs.

\subsection{Reinforcement Learning for LLM Reasoning}
Recent work has demonstrated that reasoning behaviors, rather than simply producing correct answers, are crucial for improving performance in reinforcement learning tasks~\citep{vassoyan2025ignoreklpenaltyboosting}. Research indicates that the structure of long chains of thought plays a key role in the learning process, while specific details of each reasoning step are less significant~\citep{gandhi2025cognitivebehaviorsenableselfimproving}. Critical tokens in reasoning chains serve as decision points where models frequently make errors, and modifying exploration around these tokens can reduce mistakes~\citep{wu2025rlvrworldtrainingworldmodels}.

\textbf{Proximal Policy Optimization (PPO)}~\citep{schulman2017proximalpolicyoptimizationalgorithms} is a foundational policy gradient method that constrains policy updates to ensure stable training. \textbf{Group Relative Policy Optimization (GRPO)}~\citep{shao2024deepseekmathpushinglimitsmathematical} extends PPO by computing advantages relative to groups of sampled outputs, making it particularly suitable for LLM fine-tuning. 
Recent refinements, VAPO~\citep{yue2025vapoefficientreliablereinforcement}, GSPO~\citep{zheng2025groupsequencepolicyoptimization}, DAPO~\citep{yu2025dapoopensourcellmreinforcement}, and Dr. GRPO~\citep{liu2025understandingr1zeroliketrainingcritical}, have further optimized reward design and advantage estimation techniques within the GRPO framework.

While these approaches focus primarily on reward optimization and policy updates, they typically treat reasoning as a forward generation process. In contrast, our work explicitly incorporates self-reflection and correction mechanisms during the reinforcement learning training phase, enabling models to iteratively refine their reasoning before producing final answers.

\subsection{Self-correction in LLMs}
Self-correction, also referred to as self-reflection or introspection, is a strategy in which language models analyze their own reasoning to identify and correct potential mistakes~\citep{an2024learningmistakesmakesllm}. This paradigm has gained momentum as a means to boost multi-step reasoning and problem-solving performance, especially in domains such as arithmetic, commonsense reasoning, and question answering~\citep{NEURIPS2022_9d560961}. 

Typically, self-reflection involves generating an initial answer, producing natural language feedback to critique that answer, and then refining the response based on this critique~\citep{NEURIPS2023_91edff07}. This process can be applied iteratively, often using the same model to both generate and evaluate solutions, and may include modules such as memory buffers or explicit meta-instruction guides~\citep{lin2025criticaltokensmattertokenlevel}. Recent work has explored token-level reflection mechanisms and early-exit strategies to improve efficiency~\citep{jiang2025flashthinkearlyexitmethod, wang2025critiquefinetuninglearningcritique}.

However, most existing self-reflection approaches operate primarily at inference time, requiring multiple forward passes or external verification. Our work differs by integrating reflection and correction directly into the reinforcement learning training process, allowing models to learn when and how to self-correct through the reward signal itself, rather than relying solely on verification or iterative refinement at test time.

\section{Conclusion}
We introduce ScRPO, a method emulating student learning via self-correction on errors. Experiments across five benchmarks show that ScRPO significantly outperforms baselines, boosting DeepSeek-R1-Distill-Qwen-1.5B by 6.0 percentage points (to 64.8\%) and the 7B model by 3.2 points (to 77.8\%). These results validate ScRPO as an effective RL approach, demonstrating that leveraging error-correction trajectories provides denser supervision than standard outcome-based signals, thereby facilitating more efficient policy optimization.

\section*{Limitations}
Despite the effectiveness of ScRPO in enhancing reasoning capabilities, a primary limitation lies in its dependence on the model's inherent instruction-following proficiency. Since the self-correction stage explicitly requires the model to perform reflective analysis on its own errors based on prompts, the model must possess a sufficient capability to comprehend and execute these complex instructions. Consequently, smaller-scale models or base models with weak instruction adherence may struggle to perform effective self-correction directly. To address this, such models typically require a ``cold start'' phase involving supervised fine-tuning (SFT) to establish the necessary alignment before ScRPO training can be effectively applied.

\section*{Ethics Statement}
This work enhances the mathematical reasoning capabilities of LLMs through self-correction, positively contributing to AI reliability by mitigating hallucinations in logical inference. We utilize established public datasets (e.g., GSM8k, AIME) that contain no personally identifiable information (PII) or offensive content. While our reinforcement learning experiments incur computational costs, we mitigate the environmental impact by open-sourcing our training framework and code to prevent redundant reproduction efforts. We foresee no direct societal risks associated with this research.

\bibliography{custom}
\clearpage
\onecolumn
\appendix
\newpage

\section{Pseudo-code of ScRPO}
\label{app:scrpo-algorithm}

For completeness, we provide the pseudo-code of the proposed ScRPO training procedure in Algorithm~\ref{alg:scrpo}. The algorithm summarizes the overall training workflow, encompassing both the trial-and-error filtering stage and the periodic self-correction updates.

\begin{algorithm}[H]
\caption{ScRPO: Self-Correction Reinforcement Policy Optimization}
\label{alg:scrpo}
\SetAlgoLined
\KwIn{Dataset $\mathcal{D}$; policy model $\pi_\theta$; reference model $\pi_{\text{ref}}$; hyperparameters $n$ (samples per problem), $G$ (group size), $K$ (correction frequency)}
\KwOut{Optimized policy $\pi_\theta$}

\textbf{Initialize:} error pool $\mathcal{E} \leftarrow \emptyset$, iteration counter $t \leftarrow 0$\;

\While{not converged}{
    $t \leftarrow t + 1$\;
    $\mathcal{P}_{\text{filter}} \leftarrow \emptyset$\;
    
    \tcp{Stage 1: Trial-and-Error Learning with Variance-Based Filtering}
    \For{each batch of problems $\mathcal{B} \subset \mathcal{D}$}{
        \For{each problem $q \in \mathcal{B}$}{
            Sample $n$ responses $\{o_1, \ldots, o_n\} \sim \pi_\theta(\cdot \mid q)$\;
            Compute accuracy $\mathrm{acc}(q) = \frac{\#\text{correct responses}}{n}$\;
            
            \If{$0.33 < \mathrm{acc}(q) < 0.66$}{
                Add $q$ to $\mathcal{P}_{\text{filter}}$\;
                Add incorrect responses of $q$ to error pool $\mathcal{E}$\;
            }
        }
    }
    
    \tcp{Policy Update on Filtered Problems}
    \If{$\mathcal{P}_{\text{filter}} \neq \emptyset$}{
        Sample $G$ trajectories for $q \in \mathcal{P}_{\text{filter}}$ using $\pi_\theta$\;
        Compute rewards and advantages\;
        Update $\pi_\theta$ using standard GRPO objective\;
    }
    
    \tcp{Stage 2: Self-Correction Learning (every $K$ iterations)}
    \If{$t \bmod K = 0$}{
        \For{each batch from $\mathcal{E}$}{
            Construct correction prompts $q_{\text{corr}}$ from $(q, o_{\text{wrong}})$\;
            Sample $G$ correction trajectories $\{o_1, \ldots, o_G\} \sim \pi_{\theta_{\text{old}}}(\cdot \mid q_{\text{corr}})$\;
            
            \For{each trajectory $o_j$}{
                Evaluate final-answer correctness to obtain reward $r_j$\;
                Identify tokens between \texttt{Analysis} and \texttt{Corrected Solution}\;
                Construct mask $m_{j,i} \in \{0,1\}$ for analysis tokens\;
            }
            
            Compute $\mathcal{L}_{\text{ScRPO}}$ using clipped PPO on masked tokens\;
            Update policy parameters $\theta$\;
        }
    }
}
\Return $\pi_\theta$
\end{algorithm}

\section{Detailed Formulation of GRPO}
\label{app:grpo_details}

In this section, we provide the comprehensive mathematical formulation of Group Relative Policy Optimization (GRPO) discussed in Section~\ref{sec:preliminary}.

Group Relative Policy Optimization (GRPO)~\citep{shao2024deepseekmathpushinglimitsmathematical} represents a specialized policy gradient approach tailored for the fine-tuning of LLMs. This method distinguishes itself from conventional Proximal Policy Optimization (PPO) by calculating advantages through comparison with a group of sampled outputs, making it particularly suitable for scenarios where a prompt can yield multiple distinct valid responses.

\textbf{Problem Formulation.} We consider a language model $\pi_\theta$ with parameters $\theta$, which produces outputs $o$ given input prompts $q$. For any given prompt $q$, the policy $\pi_\theta$ generates a group of $G$ responses denoted as $\{o_1, o_2, \ldots, o_G\}$. A reward score $r(q, o_i)$ is assigned to each response according to task-dependent evaluation metrics (e.g., accuracy).

\textbf{Computing Advantages.} The advantage metric quantifies the degree to which an individual response outperforms the group's mean performance:
\begin{equation}
    \hat{A}_i = \frac{r(q, o_i) - \bar{r}}{\sigma}, \quad \text{where } \bar{r} = \frac{1}{G}\sum_{j=1}^{G} r(q, o_j)
\end{equation}
Note that standardization (dividing by standard deviation $\sigma$) is often applied in practice to stabilize training.

\textbf{Objective Function.} The optimization objective for GRPO is formulated as:
\begin{equation}
\label{eq:GRPO-obj}
\begin{aligned}
    \mathcal{J}_{\text{GRPO}}(\theta) = \mathbb{E}_{q \sim P(Q), \{o_i\}_{i=1}^G \sim \pi_{\theta_{old}}} \Bigg[ &\frac{1}{G} \sum_{i=1}^G \frac{1}{|o_i|} \sum_{t=1}^{|o_i|} \min \bigg( \\
    &\rho_{i,t} \hat{A}_{i,t}, \quad \text{clip}(\rho_{i,t}, 1 - \epsilon, 1 + \epsilon) \hat{A}_{i,t} \bigg) \\
    &- \beta \mathbb{D}_{KL}(\pi_{\theta} || \pi_{ref}) \Bigg]
\end{aligned}
\end{equation}

where the notation is defined as follows:
\begin{itemize}
    \item $o_{i,t}$ represents the token at position $t$ within the $i$-th generated output.
    \item $\rho_{i,t} = \frac{\pi_\theta(o_{i,t} | q, o_{i,<t})}{\pi_{\theta_{old}}(o_{i,t} | q, o_{i,<t})}$ is the importance weight for token $o_{i,t}$.
    \item $\hat{A}_{i,t}$ denotes the advantage value. In GRPO, the same advantage score $\hat{A}_{i}$ is typically assigned to every token $t$ in response $o_i$.
    \item $\epsilon$ is the clipping threshold (e.g., 0.2).
    \item $\beta$ controls the strength of the KL divergence penalty.
    \item $\pi_{ref}$ is the reference policy (usually the SFT model).
\end{itemize}

We estimate the KL divergence term using the token-level approximation:
\begin{equation}
    \mathbb{D}_{KL}\left[\pi_{\theta} || \pi_{ref}\right] \approx \log \frac{\pi_{\theta}(o_{i,t}|q,o_{i,<t})}{\pi_{ref}(o_{i,t}|q,o_{i,<t})}
\end{equation}
or the unbiased estimator used in PPO:
\begin{equation}
    \text{KL} \approx \frac{\pi_{ref}}{\pi_{\theta}} - \log\frac{\pi_{ref}}{\pi_{\theta}} - 1
\label{eq:kl_approx_tokenwise}
\end{equation}
This term ensures the policy does not deviate excessively from the reference model, maintaining training stability.

\section{Benchmarks and Training Details}
\label{sec:benchmarks_and_training_details}  

\subsection{Benchmarks}
We evaluate mathematical reasoning performance across five distinct datasets ranging from elementary to Olympiad-level difficulty: 
\textbf{GSM8k}~\citep{cobbe2021trainingverifierssolvemath} contains grade school math word problems requiring multi-step reasoning. 
\textbf{MATH-500}~\citep{lightman2023letsverifystepstep} comprises 500 challenging competition problems spanning algebra, number theory, and geometry. 
\textbf{AIME-2024} consists of problems from the 2024 American Invitational Mathematics Examination. 
\textbf{AMC} includes multiple-choice problems from AMC 10 and AMC 12, covering high school mathematics curriculum. 
\textbf{OlympiadBench}~\citep{he2024olympiadbenchchallengingbenchmarkpromoting} (referred to as Olympiad) features problems from advanced competitions such as IMO and USAMO, requiring deep mathematical insight and rigorous proofs. 
We report the top-1 accuracy (Acc) for all benchmarks.

\subsection{Training Details}
\textbf{Dataset and Framework.} We utilize a mixed-difficulty dataset constructed by combining MATH and DAPO-MATH~\citep{han2025modelsthoughtenoughtraining}. We filter out all Chinese-language instances, resulting in approximately 14,000 examples. We explicitly verify that there is no problem-level overlap between this training set and any of our evaluation benchmarks. All models are trained using the VeRL framework~\citep{Sheng_2025}.

\textbf{Hyperparameters.} To ensure consistent optimization dynamics, both the trial-and-error learning stage and the self-correction learning stage share the same hyperparameter configuration. We employ the AdamW optimizer~\citep{loshchilov2019decoupledweightdecayregularization} with a learning rate of $1 \times 10^{-6}$. The prompt batch size is set to $B=128$, with $G=12$ sampled responses generated per prompt. For generation, we enforce a maximum length of 10,000 tokens, a temperature of 0.6, and a top-$p$ of 0.95. The clipping thresholds are fixed at $\epsilon_{\text{high}}=0.27$ and $\epsilon_{\text{low}}=0.2$. For the Variance-Based Filter, the variance threshold corresponds to the accuracy range $(0.33, 0.66)$ as defined in Eq.~(6).

\textbf{Baselines and Compute.} To ensure a fair comparison, we ensure that the baseline methods are trained for a sufficient number of steps to reach convergence, with total computational duration comparable to that of ScRPO. All reinforcement-learning-related baseline methods use the same parameter settings as ScRPO. All experiments were conducted on $8 \times$ NVIDIA H200 GPUs equipped with Intel Xeon Platinum 8558 CPUs.

\subsection{Evaluation Details}
We report the \textbf{average accuracy} (pass@1) across all benchmarks. To reduce variance in evaluation, we sample multiple responses per problem and calculate the average correctness: we use 16 samples for AIME-2024 and AMC, and 8 samples for GSM8k, MATH-500, and Olympiad.
To maintain consistency, all baselines and ScRPO are evaluated using identical prompts and inference parameters: a temperature of 0.6, top-$p$ of 0.95, and a maximum new token limit of 32,768.

\section{Additional Ablation Studies}
\label{sec:appendix_ablation}

\subsection{Ablation Study on Variance-Based Filter}
\label{app:ablation_vbf}
To investigate the impact of the Variance-Based Filter (VBF) and validate our theoretical derivation of the difficulty interval, we conducted a comprehensive ablation study across two model scales: DeepSeek-R1-Distill-Qwen-1.5B and 7B. We evaluated the models on five mathematical reasoning benchmarks (AIME24, AMC, Olympiad, GSM8k, and MATH-500) by adjusting the variance threshold from 0 (using all data) to 0.25 (retaining only the highest-uncertainty samples).

\begin{figure*}[t]
    \centering
    \includegraphics[width=1\linewidth]{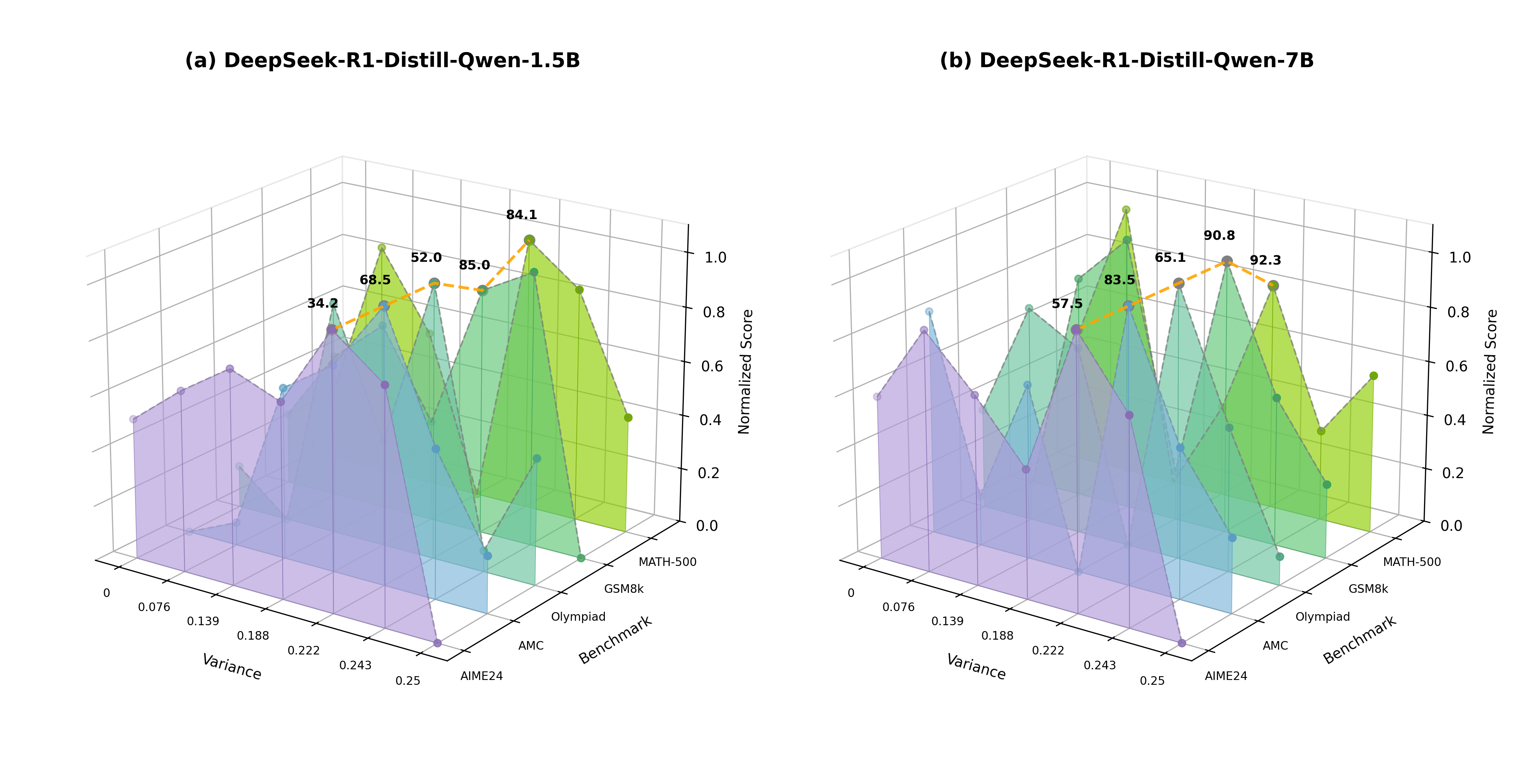} 
    \caption{Ablation study of the Variance-Based Filter on DeepSeek-R1-Distill-Qwen-1.5B and 7B. The x-axis denotes the variance threshold. The highlighted regions indicate that the threshold of 0.222 consistently achieves optimal or near-optimal accuracy across diverse benchmarks.}
    \label{fig:variance_ablation}
\end{figure*}

The experimental results, visualized in Figure~\ref{fig:variance_ablation}, reveal a consistent inverted U-shaped trend across both model scales, where performance peaks around the variance threshold of 0.222. This empirical evidence strongly supports our selection of the accuracy interval $(0.33, 0.66)$.

\paragraph{Analysis on 1.5B Model.}
For the smaller 1.5B parameter model, the threshold of 0.222 proves to be critical for efficient learning.
\begin{itemize}
    \item \textbf{Peak Performance:} The model achieves its highest scores at variance 0.222 on three major benchmarks: \textbf{AIME24} (34.2\%), \textbf{AMC} (68.5\%), \textbf{Olympiad} (52.0\%), and \textbf{MATH-500} (84.1\%). Notably, on the difficult AIME24 dataset, performance improves significantly from 32.2\% (baseline) to 34.2\% at this threshold.
    \item \textbf{Stability:} While results on GSM8k show slight fluctuations, the score at 0.222 (85.0\%) remains extremely competitive, deviating less than 0.2\% from the global maximum. This indicates that the 0.222 threshold provides a robust balance between sample difficulty and training stability for smaller models.
\end{itemize}

\paragraph{Analysis on 7B Model.}
The effectiveness of the VBF is even more pronounced in the larger 7B model, demonstrating scalability.
\begin{itemize}
    \item \textbf{Consistent Dominance:} The threshold of 0.222 yields the global maximum performance on \textbf{four out of five} benchmarks: \textbf{AIME24} (57.5\%), \textbf{AMC} (83.5\%), \textbf{Olympiad} (65.1\%), and \textbf{GSM8k} (90.8\%).
    \item \textbf{Clear Advantage:} On the Olympiad benchmark, filtering with variance 0.222 boosts accuracy by 1.2\% compared to the baseline (63.9\% $\rightarrow$ 65.1\%), confirming that focusing on ``learning boundary'' samples is more effective than training on the entire dataset or solely on the hardest samples.
\end{itemize}

Across both 1.5B and 7B scales, the variance threshold of $\approx 0.222$ consistently outperforms both the unfiltered baseline (variance 0) and the maximum variance setting (0.25). This confirms that problems with intermediate accuracy (between 0.33 and 0.66) offer optimal information gain for ScRPO, avoiding the noise of overly difficult problems while skipping the redundancy of trivial ones.

\subsection{Ablation Study on Ratio Clipping}
\label{app:ablation_rc}
The experimental results, shown in Table~\ref{table:clip_ratio_comparison}, demonstrate that the choice of clip ratio significantly influences the performance of ScRPO. Across both model scales, a larger upper clip ratio of 0.27 consistently yields superior results compared to the stricter constraint of 0.2.

\begin{itemize}
    \item For the \textbf{1.5B model}, reducing the upper clip ratio from 0.27 to 0.2 leads to a broad decline in performance, with the average accuracy dropping by 1.1 points. This negative impact is particularly pronounced on challenging benchmarks such as Olympiad (-1.8) and AIME24 (-1.5).
    
    \item A similar trend is observed in the \textbf{7B model}, where tightening the clip ratio to (0.2, 0.2) results in a 0.8 point decrease in average accuracy compared to the (0.2, 0.27) configuration.
\end{itemize}

These findings suggest that a moderately relaxed upper clipping bound (0.27) allows for more effective policy optimization in mathematical reasoning tasks. Our analysis suggests that an overly strict clipping threshold constrains the model's learning capacity by limiting the magnitude of policy updates. In contrast, a higher clipping coefficient affords the model greater flexibility to deviate from the reference policy, thereby maintaining a crucial level of exploration during the training process. This exploratory capability effectively prevents the optimization trajectory from converging prematurely to local optima, enabling the model to navigate the complex solution space of mathematical problems more robustly.

\begin{table*}[t]
\centering
\footnotesize
\caption{Comparison of ScRPO and GRPO methods with different clip ratios on mathematical reasoning benchmarks.}
\label{table:clip_ratio_comparison}
\setlength\tabcolsep{5pt} 
\renewcommand{\arraystretch}{1.1}
\begin{tabular}{@{}lllllll@{}} 
\toprule
\textbf{Method} & \textbf{AIME24} & \textbf{AMC} & \textbf{Olympiad} & \textbf{GSM8k} & \textbf{MATH-500} & \textbf{Average} \\ 
\midrule
\\[-0.8em]
\multicolumn{7}{l}{{\cellcolor[rgb]{0.957,0.957,0.957}}\textit{\textbf{DeepSeek-R1-Distill-Qwen-1.5B}}} \\
\\[-0.8em]
\textit{ScRPO (clip ratio 0.2, 0.27)} & 34.2 & 68.5 & 52.0 & 85.0 & 84.1 & 64.8 \\
\textit{ScRPO (clip ratio 0.2, 0.2)} & 32.7 {\scriptsize\textcolor{blue}{(-1.5)}} & 67.1 {\scriptsize\textcolor{blue}{(-1.4)}} & 50.2 {\scriptsize\textcolor{blue}{(-1.8)}} & 84.4 {\scriptsize\textcolor{blue}{(-0.6)}} & 83.9 {\scriptsize\textcolor{blue}{(-0.2)}} & 63.7 {\scriptsize\textcolor{blue}{(-1.1)}} \\
\textit{GRPO (clip ratio 0.2, 0.27)} & 31.1 {\scriptsize\textcolor{blue}{(-3.1)}} & 66.8 {\scriptsize\textcolor{blue}{(-1.7)}} & 49.9 {\scriptsize\textcolor{blue}{(-2.1)}} & 84.7 {\scriptsize\textcolor{blue}{(-0.3)}} & 83.6 {\scriptsize\textcolor{blue}{(-0.5)}} & 63.2 {\scriptsize\textcolor{blue}{(-1.6)}} \\
\\[-0.3em]
\midrule
\\[-0.8em]
\multicolumn{7}{l}{{\cellcolor[rgb]{0.957,0.957,0.957}}\textit{\textbf{DeepSeek-R1-Distill-Qwen-7B}}} \\
\\[-0.8em]
\textit{ScRPO (clip ratio 0.2, 0.27)} & 57.5 & 83.5 & 65.1 & 90.8 & 92.3 & 77.8 \\
\textit{ScRPO (clip ratio 0.2, 0.2)} & 56.0 {\scriptsize\textcolor{blue}{(-1.5)}} & 82.6 {\scriptsize\textcolor{blue}{(-0.9)}} & 64.5 {\scriptsize\textcolor{blue}{(-0.6)}} & 90.2 {\scriptsize\textcolor{blue}{(-0.6)}} & 91.7 {\scriptsize\textcolor{blue}{(-0.6)}} & 77.0 {\scriptsize\textcolor{blue}{(-0.8)}} \\
\textit{GRPO (clip ratio 0.2, 0.27)} & 55.3 {\scriptsize\textcolor{blue}{(-2.2)}} & 82.7 {\scriptsize\textcolor{blue}{(-0.8)}} & 64.7 {\scriptsize\textcolor{blue}{(-0.4)}} & 89.3 {\scriptsize\textcolor{blue}{(-1.5)}} & 92.0 {\scriptsize\textcolor{blue}{(-0.3)}} & 76.8 {\scriptsize\textcolor{blue}{(-1.0)}} \\
\bottomrule
\end{tabular}
\end{table*}

\subsection{Ablation Study on Self-correction Activation Frequency}
\label{app:ablation_f}

\begin{figure*}[t]
    \centering
    \includegraphics[width=1\linewidth]{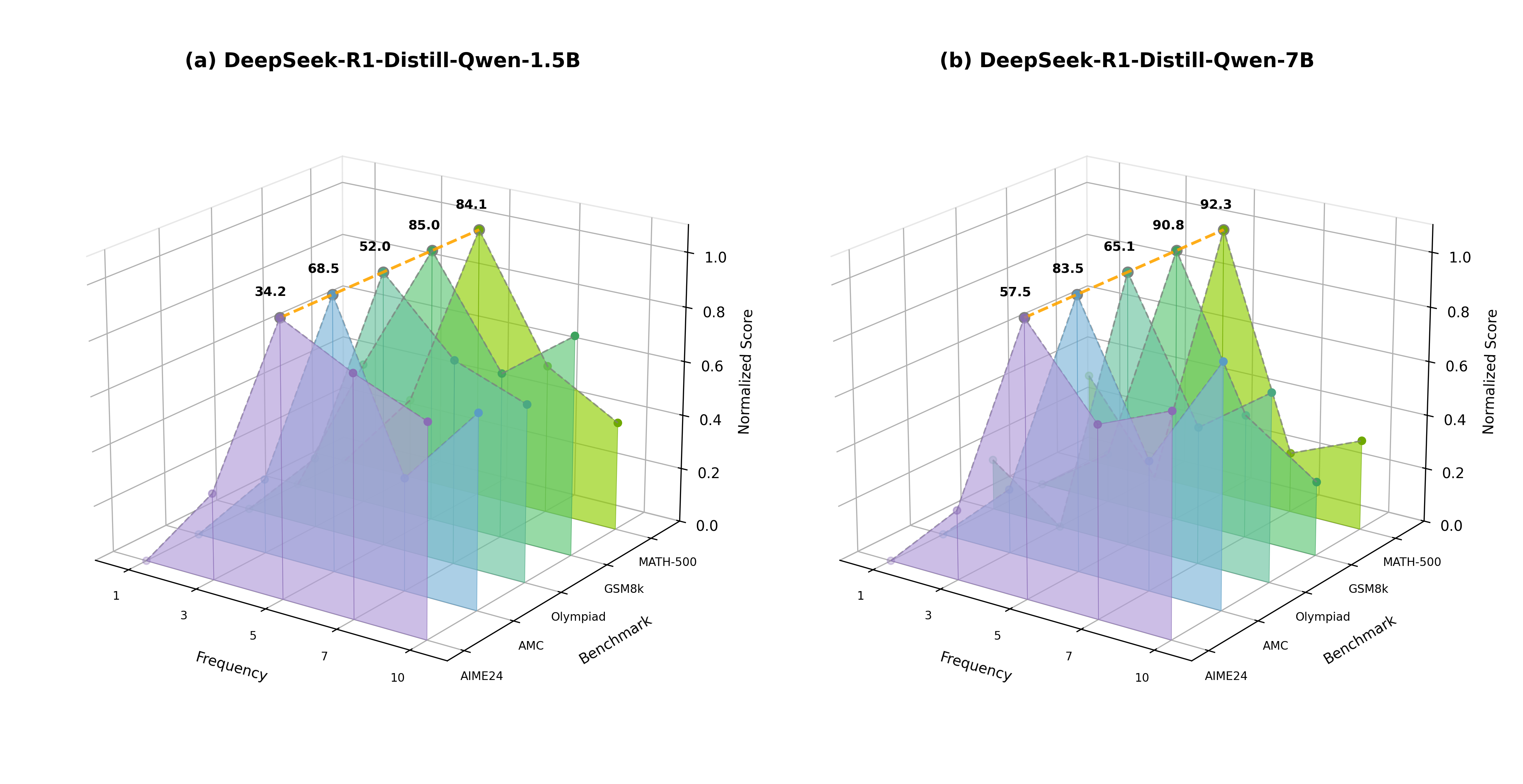} 
    \caption{Ablation study of Self-correction Activation Frequency on DeepSeek-R1-Distill-Qwen-1.5B and 7B. The x-axis denotes the frequency of Self-correction activation. }
    \label{fig:frequency_ablation}
\end{figure*}

We investigated the impact of the alternation frequency between the trial-and-error learning stage and the self-correction learning stage. By varying the interval $k$ (the number of training steps in the first stage before triggering the second stage) within the set $\{1, 3, 5, 7, 10\}$, we observed the performance trends on both the 1.5B and 7B models.

The experimental results, shown in Figure~\ref{fig:frequency_ablation}, demonstrate a distinct concave trajectory. The model performance peaks consistently at $k=5$, achieving the highest average accuracy of 64.8\% (1.5B) and 77.8\% (7B). The performance drops noticeably when the interval is either too short or too long. We attribute these phenomena to the following mechanisms:

\begin{itemize}
    \item \textbf{Instability caused by frequent switching ($k < 5$):} When the self-correction stage is triggered too frequently (e.g., $k=1$ or $k=3$), the training process exhibits significant instability. We hypothesize that the rapid alternation between the exploratory objective of the trial-and-error stage and the reflective objective of the self-correction stage prevents the optimizer from establishing a consistent gradient direction. This leads to unstable parameter updates, hindering the model's ability to effectively consolidate knowledge from either stage.

    \item \textbf{Model bias caused by delayed correction ($k > 5$):} Conversely, extending the interval (e.g., $k=10$) results in the excessive accumulation of negative samples in the error pool. Consequently, when the self-correction stage is finally activated, the model is forced to process a disproportionately large volume of correction tasks at once. This intensive and prolonged focus on error rectification causes the model's learning trajectory to drift, effectively introducing a bias where the model overfits to the specific mechanics of self-correction at the expense of its general reasoning capabilities.
\end{itemize}

Based on these findings, we conclude that an interval of $k=5$ provides the optimal trade-off, maintaining training stability while preventing the accumulation of excessive error noise. Therefore, we adopt $k=5$ as the default setting for our ScRPO framework.

\end{document}